# Fine-Tuning DialoGPT on Common Diseases in Rural Nepal for Medical Conversations

Birat Poudel[1], Satyam Ghimire[2], Er. Prakash Chandra Prasad[3]

[1]Department of Electronics and Computer Engineering, Thapathali Campus, IOE, TU, E-mail:poudel.birat25@gmail.com
[2]Department of Electronics and Computer Engineering, Thapathali Campus, IOE, TU, E-mail:satyamghimirestar@gmail.com
[3]Department of Electronics and Computer Engineering, Pulchowk Engineering Campus, IOE, TU, Email: prakash.chandra@pcampus.edu.np

*Abstract*— **Conversational agents are increasingly being explored to support healthcare delivery, particularly in resource-constrained settings such as rural Nepal. Large-scale conversational models typically rely on internet connectivity and cloud infrastructure, which may not be accessible in rural areas. In this study, we fine-tuned DialoGPT, a lightweight generative dialogue model that can operate offline, on a synthetically constructed dataset of doctor–patient interactions covering ten common diseases prevalent in rural Nepal, including common cold, seasonal fever, diarrhea, typhoid fever, gastritis, food poisoning, malaria, dengue fever, tuberculosis, and pneumonia. Despite being trained on a limited, domain-specific dataset, the fine-tuned model produced coherent, contextually relevant, and medically appropriate responses, demonstrating an understanding of symptoms, disease context, and empathetic communication. These results highlight the adaptability of compact, offline-capable dialogue models and the effectiveness of targeted datasets for domain adaptation in low-resource healthcare environments, offering promising directions for future rural medical conversational AI.**

## Introduction

Rural healthcare in Nepal faces significant challenges due to limited medical infrastructure, scarcity of trained personnel, and constrained access to reliable information. Patients in these areas frequently seek guidance on common illnesses such as the common cold, seasonal fever, diarrhea, typhoid fever, gastritis, food poisoning, malaria, dengue fever, tuberculosis, and pneumonia. Traditionally, these queries are addressed by local healthcare workers or visiting doctors, but the increasing population and remoteness of some communities make timely, personalized support difficult to provide. This gap highlights the need for scalable, automated solutions that can assist patients with accurate and contextually relevant medical advice.

Pre-trained dialogue models such as DialoGPT (dialogue generative pre-trained transformer) can generate coherent responses in open-domain settings, but their performance in specialized medical contexts is limited without domain-specific adaptation. Moreover, rural healthcare contexts pose additional challenges: internet connectivity and electricity may be unreliable, making cloud-based solutions difficult to deploy. Lightweight, offline-capable models offer a practical alternative for such low-resource environments.

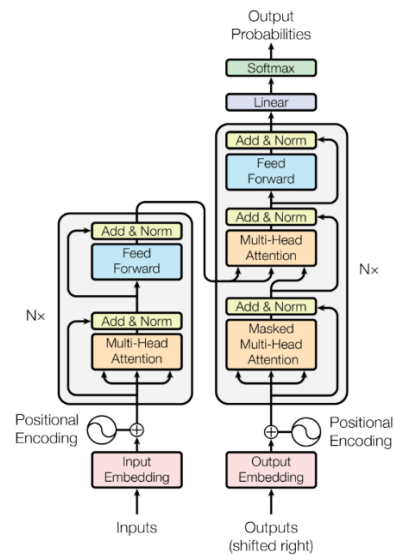

Fig. 1 The Transformer - model architecture

Our approach to overcoming the scarcity of medical dialogue data with rural healthcare context is the generation of synthetic datasets. By constructing question–answer pairs that reflect typical doctor–patient interactions for common rural illnesses, it is possible to simulate realistic medical conversations. Synthetic data generation is cost-effective, scalable, and enables the development of domain-adapted dialogue systems even in settings with limited access to annotated corpora.

In this study, we fine-tune DialoGPT on a synthetically constructed dataset representing medical conversations about ten common diseases in rural Nepal. The goal is to enable the model to generate empathetic, medically accurate, and context-aware responses while remaining suitable for offline deployment. Our work demonstrates the feasibility of adapting pre-trained conversational models to low-resource healthcare environments, providing a pathway toward accessible and scalable medical dialogue systems in rural communities.

## Related Works

Recent progress in medical conversational AI systems focuses on the generation of specialized medical datasets supported by domain-specific modeling techniques. A widely used foundation model for conversation is DialoGPT, a large-scale generative pre-trained transformer trained on 147 million Reddit conversational exchanges; DialoGPT has become a common starting point for fine-tuning conversational systems because of its strong conversational fluency and context modeling [1].

At the same time, a number of large medical dialogue corpora have been released that provide the domain coverage necessary for clinical fine-tuning and evaluation. MedDialog collected hundreds of thousands to millions of doctor-patient exchanges in English and Chinese and has been extensively used for training and transfer learning in medical dialogue research [2]. Complementing those broad collections, MedDG provides an entity-centric dataset for gastrointestinal consultations with detailed entity annotations (symptoms, tests, medicines), and shows that explicitly modeling entities helps produce more medically relevant responses [3].

Methodological advances in medical dialogue modelling reflect the specific challenges of clinical conversations: limited labeled data, the need for explicit state tracking, and the requirement for factual correctness, etc. For instance, VRBot introduces semi-supervised variational reasoning with latent variables representing patient state and physician actions to improve reasoning under scarce labels [4]. ReMeDi offers multi-domain, multi-service dialogues with fine-grained annotations designed to benchmark task-oriented medical systems across services (e.g., diagnosis, triage, prescription, etc.) [5].

For downstream clinical tasks that draw directly on conversation context, datasets such as DialMed (dialogue-based medication recommendation) demonstrate how dialogue data can be used for concrete clinical decision tasks [6]. The Dual-Flow / DFMed family of models argues for modeling both entity transitions and dialogue-act flows to better capture how medical dialogues evolve turn by turn [7]. Another active direction is knowledge-grounded generation, where medical knowledge graphs or augmented knowledge sources are integrated with generative models to reduce hallucination and boost factuality (an approach shown to be effective in recent work on augmented-graph grounding). [8] Parallel work on domain-specific pretraining has shown clear benefits for medical language tasks. Generative biomedical pretraining such as BioGPT improves biomedical text generation and QA relative to general models by pretraining on PubMed literature [9].

On the encoder side, models like ClinicalBERT and BioBERT, pretrained respectively on clinical notes and biomedical corpora, consistently improve downstream clinical NLP tasks (entity extraction, relation extraction, predictive tasks, etc.) and are commonly used as components or baselines in medical dialogue research [10] [11]. Finally, several surveys and meta-analyses synthesize progress across these areas and emphasize recurring evaluation and safety challenges, particularly the need for human-centered clinical evaluation, domain adaptation to low-resource languages, and rigorous assessment of factuality and risk when models are deployed in real clinical or community settings [12].

Extending beyond these foundational works, recent datasets such as RealMedDial have collected real telemedical dialogues from short-video clips, enabling research on realistic doctor–patient interaction in online video settings [13]. Low-resource modelling methods such as Graph-Evolving Meta-Learning (GEML) enable medical dialogue generation by transferring diagnostic experience via evolving disease–symptom graphs, addressing situations with limited labelled dialogue data [14]. Knowledge-enhanced frameworks like the work titled Knowledge-Grounded Medical Dialogue Generation using Augmented Graphs integrate pre-trained language models with medical knowledge graphs to produce responses that are both fluent and clinically appropriate [15]. Terminology-aware frameworks such as Terminology-Aware Medical Dialogue Generation incorporate domain-specific terminology representation and recognition tasks into the generative process to bridge the gap between general language models and medical domain specificity [16].

Recent developments in multi-modal and enriched-knowledge generation frameworks include KI-MMDG: Knowledge-Infused Multi-modal Medical Dialogue Generation, which integrate visual cues (e.g., skin lesions) and conversation context together with medical knowledge graphs for improved diagnosis-relevant dialogue generation [17]. Additional recent datasets such as MediTOD provide doctor–patient dialogues with annotated turns for task-oriented dialogue, enabling systematic research on medical history taking and physician–patient interaction [18]. Knowledge-enhanced extraction works such as A Knowledge-Enhanced Two-Stage Generative Framework for Medical Dialogue Information Extraction address the challenge of extracting term-status pairs from medical dialogues using staged generative frameworks to better model relation and status inference [19]. And newer works like MedKP: Medical Dialogue with Knowledge Enhancement and Clinical Pathway Encoding integrate external knowledge modules and clinical pathway encoding to reduce hallucinations and improve domain-specific generation performance [20].

## Methodology

### A. Synthetic Dataset Generation

Due to the lack of publicly available medical dialogue datasets in the Nepali rural context, a synthetic dataset was constructed to simulate realistic doctor–patient interactions. The dataset focused on ten common diseases frequently encountered in rural Nepal: common cold, seasonal fever, diarrhea, typhoid fever, gastritis, food poisoning, malaria, dengue fever, tuberculosis, and pneumonia.

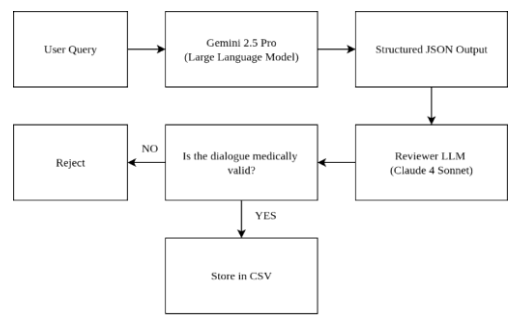

Fig. 2 Synthetic Dataset Generation Block Diagram

A pipeline was developed where Gemini 2.5 Pro generates a total of 1000 doctor–patient dialogues, 100 per each disease, in JSON format from user queries. These outputs are reviewed by Claude 4 Sonnet to assess medical validity, and only accurate dialogues are stored in a CSV file while others are rejected. The finalized dataset was later reviewed by medical professionals to ensure clinical reliability and authenticity.

Each dialogue represents typical symptom descriptions and medically appropriate advice. The dialogues emphasized:

1. Clear symptom reporting (e.g., "I have been experiencing fever and body aches for three days").
2. Providing empathetic medical responses while encouraging users to seek professional care when needed.

To maintain linguistic simplicity and accessibility, the dataset was written in English with phrasing adapted to reflect how patients in rural Nepal typically describe symptoms. The dataset was divided into training (80%) and validation (20%) splits to fine-tune and evaluate model generalization.

### B. Fine-tuning DialoGPT

We used DialoGPT-medium, a pre-trained conversational transformer model based on GPT-2 architecture, as the foundation for fine-tuning. The model was fine-tuned using the Hugging Face Transformers library. Each dialogue sample was formatted with alternating user and system turns to align with DialoGPT's expected conversational input.

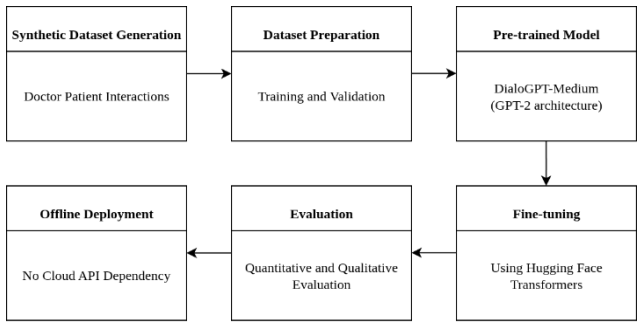

Fig. 3 Fine-tuning DialoGPT Block Diagram

The training process aimed to adapt DialoGPT's general conversational ability to domain-specific medical dialogue, improving its contextual understanding of disease-related symptoms and responses. Fine-tuning was conducted on a Google Colab T4 GPU, and the resulting model was exported for offline deployment, removing dependency on cloud APIs.

### C. Evaluation Metrics

Model evaluation was conducted using a combination of quantitative and qualitative approaches to assess both linguistic quality and medical relevance of generated responses.

**1. Quantitative Evaluation**

**Evaluation Loss:** The evaluation (validation) loss, computed as the average cross-entropy between predicted and target tokens on the validation set, was monitored throughout training. A lower loss indicates improved learning and better model generalization.

$$Loss = -\frac{1}{N}\sum_{i=1}^{N} logP(x_i \mid x_{<i}) \qquad (1)$$

where Loss is the average cross-entropy loss per token:

1. N = total number of tokens in the sequence
2. $x_i$ = the i-th target token
3. $x_{<i}$ = all previous tokens before $x_i$
4. $P(x_i \mid x_{<i})$ = predicted probability of the correct token

**Perplexity (PPL):** Perplexity was derived from the exponential of the average cross-entropy loss, serving as a measure of the model's fluency and predictive confidence. Lower perplexity corresponds to more coherent and natural responses.

$$Perplexity = e^{Loss} \qquad (2)$$

**Accuracy, Precision, Recall, and F1-Score**: During training, these metrics were computed at the token level to evaluate the model's predictive performance in the causal language modeling framework.

1. Accuracy: Represents the proportion of tokens correctly predicted by the model compared to the reference response.
2. Precision: Measures the fraction of generated tokens that were correct, helping assess how often the model produced relevant words without unnecessary additions.
3. Recall: Indicates the fraction of reference tokens the model successfully predicted, reflecting its completeness in reproducing contextually appropriate responses.
4. F1-Score: The harmonic mean of precision and recall, providing a balanced evaluation of both correctness and coverage in token prediction.

These metrics collectively assess how effectively the fine-tuned DialoGPT model learns to generate accurate, fluent, and contextually aligned responses within medical dialogues.

**2. Qualitative Evaluation**

A human evaluation comprising 100 dialogue samples was conducted by healthcare professionals to assess the overall conversational quality of the fine-tuned model. The evaluation focused on three key dimensions:

1. **Medical Appropriateness:** The accuracy, reliability, and safety of the medical information or advice provided.
2. **Empathy and Tone:** The model's ability to express understanding, reassurance, and supportive communication.
3. **Contextual Relevance:** The extent to which responses appropriately addressed the patient's described symptoms and situation.

Each dimension was rated on a 5-point Likert scale, and the average scores were calculated to determine the model's overall conversational effectiveness.

### D. System Methodology

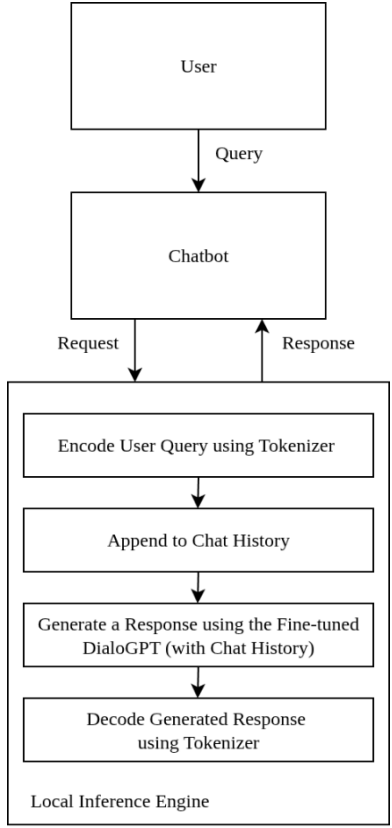

Fig. 4 System Block Diagram

The proposed system is an offline conversational chatbot designed to generate human-like responses without requiring an internet connection. The overall architecture consists of two main components: the user interface and the offline inference engine. When a user inputs a query through the chatbot interface, the system processes the request locally. First, the input text is encoded using a pretrained tokenizer, which converts the query into a sequence of numerical tokens understandable by the model. These tokens are then appended to the existing chat history to maintain conversational context across multiple interactions. The combined input sequence is passed to a fine-tuned DialoGPT model, which generates an appropriate response based on the conversation history.

The model outputs tokenized text, which is subsequently decoded back into natural language using the same tokenizer. The generated response is then displayed to the user through the chatbot interface. Since all components, including the tokenizer and the fine-tuned DialoGPT model, are locally hosted, the system can operate fully offline, ensuring data privacy, low latency, and independence from external servers.

## Results and Analysis

### A. Fine-tuning DialoGPT

DialoGPT was fine-tuned on a synthetically constructed dataset of doctor–patient dialogues covering ten common diseases prevalent in rural Nepal. The model was trained to generate contextually relevant and medically appropriate responses while remaining lightweight enough for offline deployment. Training was performed using the Causal Language Modeling (CLM) objective with cross-entropy loss.

#### 1. Perplexity Score

The model achieved a perplexity score of **5.9632**, indicating that it generates fluent and coherent responses with relatively low uncertainty in next-token prediction. Perplexity, derived as the exponential of the cross-entropy loss, provides a measure of how confidently the model predicts the next token in a sequence. Lower perplexity values suggest better language modeling performance.

#### 2. Accuracy, Precision, Recall and F1-Score

Token-level metrics were computed to assess the model's predictive ability within the domain-specific dialogue context. The results are summarized below:

TABLE 1

ACCURACY, PRECISION, RECALL, AND F1-SCORE

| Metric | Score |
|---|---|
| Accuracy | 0.1633 |
| Precision | 0.1561 |
| Recall | 0.1496 |
| F1-Score | 0.1633 |

These metrics reflect the model's ability to generate correct and contextually relevant tokens. While token-level accuracy and F1-score appear low, this is expected given the limited size of the synthetic dataset and the complexity of natural language generation in medical dialogues. Importantly, the model maintains linguistic coherence and can produce medically reasonable responses.

#### 3. Training and Evaluation Loss

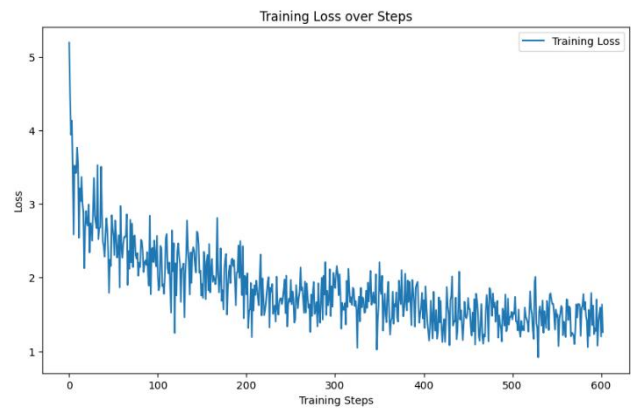

Fig. 5 Training Loss

The above plot shows the model's training loss over 600 training steps. The loss is a measure of how wrong the model's predictions are on the data it's being trained on. As the model sees more data (i.e., as training steps increase), it adjusts its internal parameters to minimize its errors, causing the loss to decrease. The fluctuations (the spiky nature of the line) are normal, as the loss is typically calculated for each small batch of data.

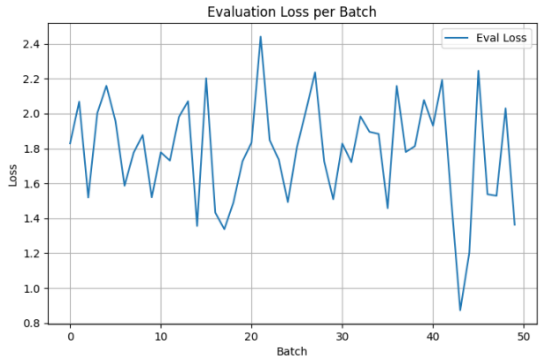

Fig. 6 Evaluation Loss

This above plot shows the model's evaluation loss, calculated for each batch of a separate evaluation dataset that the model has not been trained on. This metric is crucial for understanding how well the model generalizes to new, unseen data. The loss fluctuates significantly from batch to batch, which is common for generative conversational models like DialoGPT.

4. Hyperparameters

The following hyperparameters were used during fine-tuning DialoGPT:

TABLE 2

MODEL HYPERPARAMETERS

| Hyperparameter | Value |
| --- | --- |
| Learning Rate | 5e-5 |
| Optimizer | Adam |
| Batch Size | 4 |
| Epochs | 3 |
| Loss Function | Cross-Entropy Loss |

B. Human Evaluation

A human evaluation was conducted on 100 dialogue samples by healthcare professionals to assess conversational quality across three dimensions: medical appropriateness, empathy and tone, and contextual relevance. Each aspect was rated on a 5-point Likert scale, and average scores were calculated. The evaluation confirmed that the model generates empathetic, context-aware, and medically reasonable responses, highlighting its potential utility as a conversational assistant in rural healthcare settings.

TABLE 3

HUMAN EVALUATION METRICS

| Metric | Mean Score |
| --- | --- |
| Medical Appropriateness | 3.8 |
| Empathy and Tone | 4.1 |
| Contextual References | 3.9 |

**Conclusion**

We fine-tuned DialoGPT, a lightweight generative dialogue model, on a synthetically constructed dataset representing medical conversations about ten common diseases in rural Nepal. The fine-tuned model demonstrated the ability to generate contextually relevant, medically appropriate, and empathetic responses, despite being trained on a limited dataset. Quantitative evaluation metrics, including perplexity, token-level accuracy, precision, recall, and F1-score, alongside qualitative assessments by healthcare professionals, confirmed the model's potential to support rural healthcare delivery.

These results highlight the feasibility of deploying offline-capable conversational AI systems in low-resource environments where internet connectivity and electricity are limited. By bridging the gap between patients and healthcare information, such models can provide timely guidance, improve symptom awareness, and reduce the burden on local healthcare personnel.

**Future Enhancements**

While the results of this study are promising, several avenues exist to further enhance the model's effectiveness and reliability. Expanding the dataset to include larger and more diverse dialogues, such as varied symptom descriptions, can improve the model's coverage and robustness.

Implementing safety filters would help identify potentially unsafe or ambiguous advice, thereby improving patient safety. Multimodal extensions, such as incorporating symptom images or voice inputs, could increase accessibility for users.

Finally, conducting field testing and deployment in real rural settings would allow the collection of user feedback, refinement of dialogue strategies, and assessment of the system's impact on healthcare outcomes. By pursuing these enhancements, the system can evolve into a robust, scalable, and safe medical conversational agent, capable of delivering meaningful healthcare support in low-resource environments.

**References**

[1] S. S. M. G. Y.-C. C. C. B. X. G. J. G. J. L. a. B. D. Yizhe Zhang, "DialoGPT: Large-Scale Generative Pre-training for Conversational Response Generation," *Proceedings of the 58th Annual Meeting of the Association for Computational Linguistics: System Demonstrations,* pp. 270-278, 2019.

[2] W. Y. Z. J. Y. Y. S. W. R. Z. M. Z. J. Z. X. D. R. Z. H. F. P. Z. S. C. P. X. Guangtao Zeng, "MedDialog: Large-scale Medical Dialogue Datasets," *Proceedings of the 2020 Conference on Empirical Methods in Natural Language Processing (EMNLP),* pp. 9241-9250, 2020.

[3] J. T. Y. C. W. L. Y. Z. X. L. Wenge Liu, "MedDG: An Entity-Centric Medical Consultation Dataset for

Entity-Aware Medical Dialogue Generation," *Natural Language Processing and Chinese Computing: 11th CCF International Conference,* pp. 447-459, 2022.

[4] Z. R. P. R. Z. C. Dongdong Li, "Semi-Supervised Variational Reasoning for Medical Dialogue Generation," in *SIGIR '21: The 44th International ACM SIGIR Conference on Research and Development in Information Retrieval*, 2021.

[5] J. P. P. R. Z. R. X. X. Guojun Yan, "ReMeDi: Resources for Multi-domain, Multi-service, Medical Dialogues," *Proceedings of the 45th International ACM SIGIR Conference on Research and Development in Information Retrieval,* pp. 3013-3024, 2022.

[6] Y. H. Z. O. W. G. H. C. G. X. J. W. Zhenfeng He, "DialMed: A Dataset for Dialogue-based Medication Recommendation," *Proceedings of the 29th International Conference on Computational Linguistics,* pp. 721-733, 2022.

[7] W. H. Y. C. J. W. W. L. Kaishuai Xu, "Medical Dialogue Generation via Dual Flow Modeling (DFMed)," *Findings of the Association for Computational Linguistics: ACL 2023,* pp. 6771-6784, 2023.

[8] A. Z. N. K. B. &. A. E. Deeksha Varshney, "Knowledge-grounded medical dialogue generation using augmented graphs," *Scientific Reports,* vol. 13, p. 3310, 2023.

[9] L. S. Y. X. T. Q. S. Z. H. P. T.-Y. L. Renqian Luo, "BioGPT: Generative Pre-trained Transformer for Biomedical Text Generation and Mining," *Briefings in Bioinformatics,* vol. 23, no. 6, 2022.

[10] J. A. R. R. Kexin Huang, "ClinicalBERT: Modeling Clinical Notes and Predicting Hospital Readmission," *Preprint arXiv,* 2019.

[11] W. Y. S. K. D. K. S. K. C. H. S. J. K. Jinhyuk Lee, "BioBERT: a pre-trained biomedical language representation model for biomedical text mining," *Bioinformatics,* vol. 36, no. 4, pp. 1234-1240, 2019.

[12] N. P. Mina Valizadeh, "The AI Doctor Is In: A Survey of Task-Oriented Dialogue Systems for Healthcare Applications," *Proceedings of the 60th Annual Meeting of the Association for Computational Linguistics (Volume 1: Long Papers),* pp. 6638-6660, 2022.

[13] H. Z. J. W. X. Z. D. H. L. Z. H. L. F. M. Bo Xu, "RealMedDial: A Real Telemedical Dialogue Dataset Collected from Online Chinese Short-Video Clips," *Proceedings of the 29th International Conference on Computational Linguistics,* p. 3342–3352, 2022.

[14] P. Z. X. L. J. T. R. Z. Z. C. L. L. Shuai Lin, "Graph-Evolving Meta-Learning for Low-Resource Medical Dialogue Generation," *Proceedings of the AAAI Conference on Artificial Intelligence ,* pp. 13362-13370, 2021.

[15] A. Z. N. K. B. A. E. Deeksha Varshney, "Knowledge graph assisted end-to-end medical dialog generation," *Artificial Intelligence in Medicine,* vol. 139, 2023.

[16] H. Z. T. L. C. L. F. G. Chen Tang, "Terminology-Aware Medical Dialogue Generation," *IEEE International Conference on Acoustics, Speech and Signal Processing (ICASSP),* pp. 1-5, 2023.

[17] S. B. P. V. J. V. M. S. S. P. B. M. D. S. T. Abhisek Tiwari, "Seeing Is Believing! towards Knowledge-Infused Multi-modal Medical Dialogue Generation," *Proceedings of the 2024 Joint International Conference on Computational Linguistics, Language Resources and Evaluation (LREC-COLING 2024),* p. 14513–14523, 2024.

[18] G. S. R. J. D. D. R. M. Vishal Vivek Saley, "MediTOD: An English Dialogue Dataset for Medical History Taking with Comprehensive Annotations," *EMNLP,* 2024.

[19] Z. N. J. S. S. X. &. B. X. Zefa Hu, "A Knowledge-enhanced Two-stage Generative Framework for Medical Dialogue Information Extraction," vol. 21, pp. 153-168, 2024.

[20] X. W. Y. Z. J. Y. Jiageng Wu, "MedKP: Medical Dialogue with Knowledge Enhancement and Clinical Pathway Encoding," *arXiv:2403.06611 ,* 2024.